\documentclass[conference,compsoc]{IEEEtran}
\usepackage{tikz}
\usepackage{amssymb}
\usepackage{amstext}
\usepackage{amsmath}
\usepackage{amsthm}

\ifCLASSOPTIONcompsoc
  \usepackage[nocompress]{cite}
\else
  \usepackage{cite}
\fi
\ifCLASSINFOpdf
\else
\fi
\begin{document}
%
\title{Online Transition-Based Feature Generation for Anomaly Detection in Concurrent Data Streams}

\author{\IEEEauthorblockN{Yinzheng Zhong}
\IEEEauthorblockA{Department of Computer Science\\University of Liverpool\\Liverpool\\UK
\\Email: y.zhong10@liverpool.ac.uk
}
\and
\IEEEauthorblockN{Alexei Lisitsa}
\IEEEauthorblockA{Department of Computer Science\\University of Liverpool\\Liverpool\\UK
\\Email: a.lisitsa@liverpool.ac.uk}
}


%


\maketitle

\begin{abstract}
In this paper, we introduce the transition-based feature generator (TFGen) technique, which reads general activity data with attributes and generates step-by-step generated data. The activity data may consist of network activity from packets, system calls from processes or classified activity from surveillance cameras. TFGen processes data online and will generate data with encoded historical data for each incoming activity with high computational efficiency. The input activities may concurrently originate from distinct traces or channels. The technique aims to address issues such as domain-independent applicability, the ability to discover global process structures, the encoding of time-series data, and online processing capability.
\end{abstract}


%
\IEEEpeerreviewmaketitle

\section{Introduction}
    Anomaly detection in data analysis typically refers to the discovery of uncommon observations of patterns that differ considerably from the majority of the data and do not adhere to a well-defined concept of normal behaviour. Chandola et al. \cite{chandola2009anomaly} introduce the applications of anomaly detection in many areas such as intrusion detection, fraud detection and industrial damage detection. This paper will introduce a generic technique for extracting features from activity changes (transitions) for use in machine learning and signal processing.
    
    Our paper is based on the paper from Zhong et al. \cite{zhong2022process}, which uses a process mining related technique for network intrusion detection. The technique in \cite{zhong2022process} is inspired by process mining \cite{van2003workflow, van2011process} algorithms that discover process models from event logs. There are procedures involved in every aspect of our daily lives, from the operations of large businesses to the management of private households. In the industrial sector, one can find both the production of automobiles and the fulfilment of customer orders. The procedure or series of activities for achieving a goal is known as the process. We use a network-based intrusion detection system as an illustration of our technique in the context where a network flow of multiple packets is treated as a process.
    
    \cite{zhong2022process} introduced the feature generation algorithm and the result for intrusion detection, but did not introduce other capabilities of the algorithm. Our paper extends the technique of \cite{zhong2022process}. and explore deeper into the technique itself. This paper adds the generality that enables standardised input from applications in different domains, on top of already existing yet introduced capabilities of discovering global process structure that may aid in anomaly detection in concurrent processes, the packet-level (event-level) processing for online detection and time-series information encoding with reasonable computational complexity.
    
    An intrusion detection system (IDS) is utilised to detect and classify security policy violations and attacks. We have network-based intrusion detection system (NIDS) and host-based intrusion detection system (HIDS) depending on the purpose of the system. The NIDS is usually deployed on infrastructures like routers and switches to detects intrusions by monitoring network activities. The HIDS, on the other hand, inspects each individual system for any unauthorised file modifications, abnormal network activity, or suspicious behaviour.
    
    In the following section, we will first explore some related works that focus on intrusion detection. As \cite{zhong2022process} is closely related to the intrusion detection domain, we will understand the problem better and discuss what benefit of the algorithm from \cite{zhong2022process} provides. Then we summarise the problems in Section \ref{existing_problems} that TFGen is able to solve. The technical details of TFGen will be presented in section \ref{tech_details}, and finally we will discuss the possible applications of TFGen and some known issues of this technique.

\section{Related Work}
\label{section:related_works}
    From the detection method perspective, signature-based intrusion detection systems (SIDS) and anomaly-based intrusion detection systems (AIDS) are typically the two types of intrusion detection systems. The SIDS uses patterns to detect intrusions, or machine learning algorithms are trained with labelled data and used to classify whether an intrusion has occurred. Snort \cite{roesch1999snort} is an example of a SIDS that detects intrusions using predefined rules. Also, data mining \cite{lee1998data,borkar2019novel,ashraf2018comparative}, machine learning \cite{agarap2018neural,hsu2019toward,roshan2018adaptive,mirsky2018kitsune}, and other statistical methods \cite{vijayasarathy2011system,lee2000information,david2015ddos} exist. The machine learning models are trained with binary or multi-class data in the case of SIDS. For the example of AIDS, Zavrak and Iskefiyeli uses Autoencoder and Support Vector Machines (SVM) for anomaly detection \cite{9113298}; Abdelmoumin et al. use ensemble learning \cite{9509761} for anomaly detection in Internet of Things (IoT). The machine learning models are trained with normal data only (one-class training) in the case of AIDS.
    
    In general, SIDS checks for incoming data characteristics that are comparable to known threats; whereas AIDS analyses the deviation between incoming data and normal behaviour, and the outcome is determined based on whether the outlier score is above the threshold. Typically, the SIDS accuracy measurement is F-score, while the AIDS accuracy measurement is receiver operating characteristic (ROC) and area under curve (AUC). The ROC curve is a graph that displays how well a classification model performs across all classification thresholds. The advantage of AIDS is that is is capable of detecting zero-day attacks, however, the disadvantage of AIDS is that it normally has a higher false positive rate (FPR) than SIDS.
    
    We must note that, although some publications are proposed to be about AIDS, they do not precisely adhere to the notion of AIDS. For example, \cite{8615300} use LSTM for intrusion detection, and \cite{8903672} use machine learning for intrusion detection in industrial network. These techniques produce unreasonably high accuracy and very low FRP and claim to be anomaly based, but they are signature based. The survey \cite{khraisat2019survey} shows the concept of AIDS accurately; however, the majority of the papers to which the survey refers are not related to AIDS. We can see the same problem in other surveys \cite{maseer2021benchmarking}.
    
    From the detection speed perspective, there are two types of intrusion detection, online and offline. Online detection monitors network activity in real-time in order to detect threats as quickly as possible. Offline detection typically examines the data logged and is executed manually by the administrator or at a predetermined interval. When discussing online intrusion detection, we anticipate the response time between an attack and the activation of an alarm to be as short as possible, which is why we consider packet-level detection methods in our approach. Numerous techniques employ packet-level detection. However, some techniques, primarily those based on data mining or machine learning, do not detect packet-level intrusions. The first reason is, the commonly used popular datasets lack packet-level information, like the well-known KDD’99 and the improved NSL-KDD datasets \cite{tavallaee2009detailed}. These datasets provide statistical values at the flow level, i.e., the data is generated only after a socket is closed or timed out. For instance, the Destination Bytes feature indicates the total number of bytes transferred from the source to the destination in a single socket; obviously, this feature cannot be extracted before the socket closes or terminates. \cite{dhanabal2015study} provides details on all features included with the KDD dataset. Second, the efficiency of these systems is insufficient to support packet-level detection.
    
    There are examples of packet-level detection systems. A good example of packet-level detection is presented in \cite{mirsky2018kitsune}. Damped incremental statistics are utilised to generate packet-level feature vectors in real-time. By retaining the prior statistical value, the value can be incrementally updated with the most recent packet data. In addition to updating the previous values, the decay function devalues older data. Statistical information between rx and tx is also extracted by implementing the 2-D statistics. There also exist pattern-based packet-level IDSs examine audit trails for network data, including \cite{roesch1999snort,wespi2000intrusion}. In general, these IDSs look for specific activities or a sequence of activities, and then makes a determination based on the defined patterns. Apart from NIDS, the Variable-Length Audit Trail Pattern approach proposed by Wespi et al. \cite{wespi2000intrusion} is used for HIDS. It captures system commands such as file-open and file-close, and then maps them into sequences of characters based on the translation table. A sequence of characters will then be divided into variable-length subsequences using the Teiresias algorithm \cite{rigoutsos1998combinatorial}. The subsequences are compared with the training sequences for calculating the boundary coverage. Finally, the algorithm will determine the intrusion with the longest group of uncovered events.

\section{Existing Problems} \label{existing_problems}
    We summarise three main issues with some of the techniques used for intrusion detection. The first is the lack of ability to perform online detection; the second is that some packet-level detection techniques have difficulty applying to encrypted data; third, most techniques lack the ability to detect the global process structures.
    
    Usually, packet-level systems are incapable of decrypting and analysing encrypted payload. Examples include the case-based agent \cite{schwartz2002case}, which uses case-based reasoning on packet XML data; the previously introduced Snort, which uses predefined rules to check for intrusions, and the technique \cite{wang2020deep}, which converts bytes of packets to grayscale images and then uses hierarchical network structure for classification.
    
    Some techniques use recurrent neural networks for intrusion detection, such as \cite{hwang2019lstm}, which uses LSTM to classify a time-series of raw packets. On small network devices, resources for training and running LSTM are not always available. Using flow-level data is another option, but this level of detection is not what we seek.
    
    Packet-level IDS with the use of historical information addresses the issue with encrypted data; however, we are now facing an issue that may result in poor performance for attacks such as DoS/DDoS and brute force attacks. These attacks are possible with a high number of concurrent connections. Theoretically, each connection may appear completely normal; therefore, the attack cannot be identified if we focus on the information provided by a single connection. These concurrent connections may be identical to the previously identified data, and a signature-based IDS may have good performance; however, any changes to the requests will render these IDSs ineffective.
    
    Process mining is intended for business process model discovery and analysis, and is capable of encoding the global process structure. We believe the ability to observe global process structure is essential for detecting attacks such as botnet, DoS/DDoS, and brute force. The activities are recorded in the event log that can be used for process mining in the future. An obvious problem is that the collection of these activities could take days or even weeks, and the event log is then used to determine the process model. This may be appropriate for offline intrusion detection, but it cannot be used traditionally for online anomaly detection.
    
    Online conformance checking is available in later research \cite{van2019online} but the anomaly detection is limited to conformance checking in process mining domain. Zhong and Lisitsa have done tests in \cite{zhong2022can} as a naive approach to use process mining on network data and then tried to detect anomaly with conformance checking. The results have turned out not promising. An interested reader may find further experimental results and explanations in \cite{zhong2022can}.

\section{Online Feature Generation on Concurrent Data Streams} \label{tech_details}
    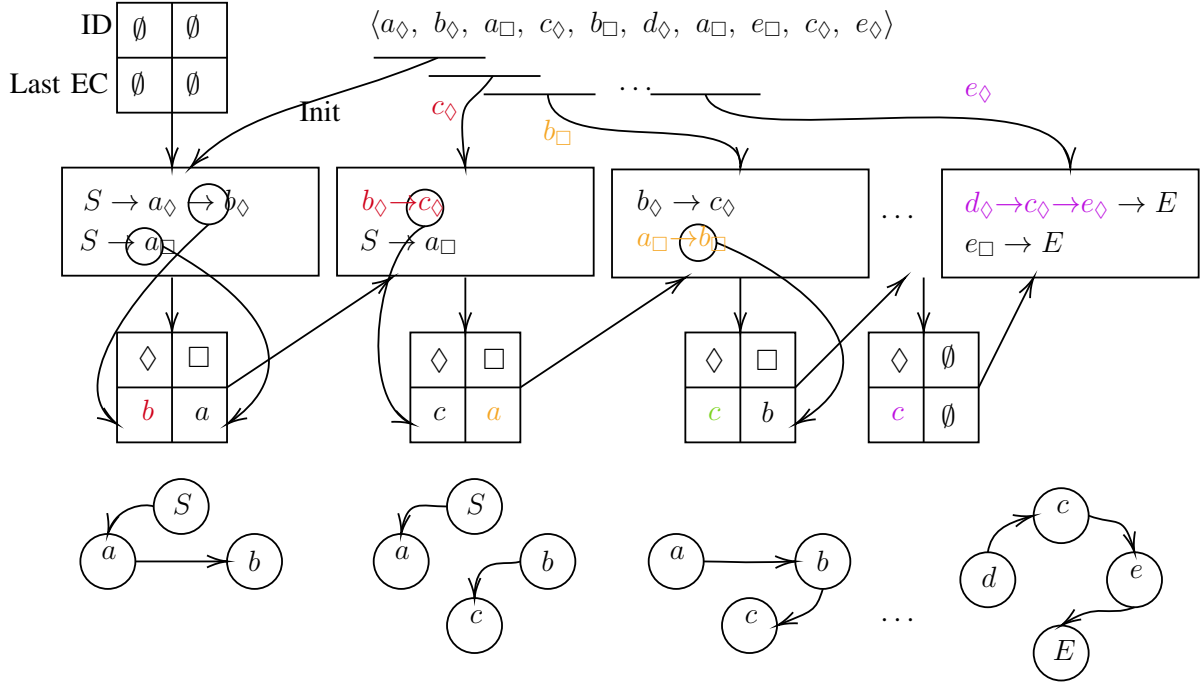
\begin{figure*}[h]
        \centering
        \large
        \tikzset{every picture/.style={line width=0.75pt}} 
        \resizebox{0.9\textwidth}{!}{%
        \begin{tikzpicture}[x=0.75pt,y=0.75pt,yscale=-1,xscale=1]
        
        \draw   (39,130) -- (179,130) -- (179,189) -- (39,189) -- cycle ;
        \draw   (189,130) -- (329,130) -- (329,189) -- (189,189) -- cycle ;
        \draw   (339,131) -- (479,131) -- (479,190) -- (339,190) -- cycle ;
        \draw   (519,131) -- (659,131) -- (659,190) -- (519,190) -- cycle ;
        
        \draw    (209,70) -- (270,70) ;
        \draw    (244,70) .. controls (182.93,95.61) and (140.29,96.96) .. (110.36,128.54) ;
        \draw [shift={(109,130)}, rotate = 312.27] [color={rgb, 255:red, 0; green, 0; blue, 0 }  ][line width=0.75]    (10.93,-3.29) .. controls (6.95,-1.4) and (3.31,-0.3) .. (0,0) .. controls (3.31,0.3) and (6.95,1.4) .. (10.93,3.29)   ;
        \draw    (239,80) -- (300,80) ;
        \draw    (274,80) .. controls (260.21,99.7) and (254.18,87.38) .. (258.78,128.1) ;
        \draw [shift={(259,130)}, rotate = 263.37] [color={rgb, 255:red, 0; green, 0; blue, 0 }  ][line width=0.75]    (10.93,-3.29) .. controls (6.95,-1.4) and (3.31,-0.3) .. (0,0) .. controls (3.31,0.3) and (6.95,1.4) .. (10.93,3.29)   ;
        \draw    (269,90) -- (330,90) ;
        \draw    (304,90) .. controls (305.96,125.28) and (407.8,97.17) .. (409.05,128.04) ;
        \draw [shift={(409,130)}, rotate = 275.04] [color={rgb, 255:red, 0; green, 0; blue, 0 }  ][line width=0.75]    (10.93,-3.29) .. controls (6.95,-1.4) and (3.31,-0.3) .. (0,0) .. controls (3.31,0.3) and (6.95,1.4) .. (10.93,3.29)   ;
        \draw    (360,90) -- (420,90) ;
        \draw    (390,90) .. controls (399.9,124.65) and (583.28,76.97) .. (588.88,128.41) ;
        \draw [shift={(589,130)}, rotate = 267.88] [color={rgb, 255:red, 0; green, 0; blue, 0 }  ][line width=0.75]    (10.93,-3.29) .. controls (6.95,-1.4) and (3.31,-0.3) .. (0,0) .. controls (3.31,0.3) and (6.95,1.4) .. (10.93,3.29)   ;
        \draw  [draw opacity=0] (69,220) -- (129,220) -- (129,280) -- (69,280) -- cycle ; \draw   (99,220) -- (99,280) ; \draw   (69,250) -- (129,250) ; \draw   (69,220) -- (129,220) -- (129,280) -- (69,280) -- cycle ;
        \draw    (99,190) -- (99,218) ;
        \draw [shift={(99,220)}, rotate = 270] [color={rgb, 255:red, 0; green, 0; blue, 0 }  ][line width=0.75]    (10.93,-3.29) .. controls (6.95,-1.4) and (3.31,-0.3) .. (0,0) .. controls (3.31,0.3) and (6.95,1.4) .. (10.93,3.29)   ;
        \draw    (129,250) -- (217.34,191.11) ;
        \draw [shift={(219,190)}, rotate = 146.31] [color={rgb, 255:red, 0; green, 0; blue, 0 }  ][line width=0.75]    (10.93,-3.29) .. controls (6.95,-1.4) and (3.31,-0.3) .. (0,0) .. controls (3.31,0.3) and (6.95,1.4) .. (10.93,3.29)   ;
        \draw  [draw opacity=0] (69,40) -- (129,40) -- (129,100) -- (69,100) -- cycle ; \draw   (99,40) -- (99,100) ; \draw   (69,70) -- (129,70) ; \draw   (69,40) -- (129,40) -- (129,100) -- (69,100) -- cycle ;
        \draw    (99,100) -- (99,128) ;
        \draw [shift={(99,130)}, rotate = 270] [color={rgb, 255:red, 0; green, 0; blue, 0 }  ][line width=0.75]    (10.93,-3.29) .. controls (6.95,-1.4) and (3.31,-0.3) .. (0,0) .. controls (3.31,0.3) and (6.95,1.4) .. (10.93,3.29)   ;
        \draw  [draw opacity=0] (229,220) -- (289,220) -- (289,280) -- (229,280) -- cycle ; \draw   (259,220) -- (259,280) ; \draw   (229,250) -- (289,250) ; \draw   (229,220) -- (289,220) -- (289,280) -- (229,280) -- cycle ;
        \draw    (259,190) -- (259,218) ;
        \draw [shift={(259,220)}, rotate = 270] [color={rgb, 255:red, 0; green, 0; blue, 0 }  ][line width=0.75]    (10.93,-3.29) .. controls (6.95,-1.4) and (3.31,-0.3) .. (0,0) .. controls (3.31,0.3) and (6.95,1.4) .. (10.93,3.29)   ;
        \draw    (289,250) -- (377.34,191.11) ;
        \draw [shift={(379,190)}, rotate = 146.31] [color={rgb, 255:red, 0; green, 0; blue, 0 }  ][line width=0.75]    (10.93,-3.29) .. controls (6.95,-1.4) and (3.31,-0.3) .. (0,0) .. controls (3.31,0.3) and (6.95,1.4) .. (10.93,3.29)   ;
        \draw  [draw opacity=0] (379,220) -- (439,220) -- (439,280) -- (379,280) -- cycle ; \draw   (409,220) -- (409,280) ; \draw   (379,250) -- (439,250) ; \draw   (379,220) -- (439,220) -- (439,280) -- (379,280) -- cycle ;
        \draw    (409,190) -- (409,218) ;
        \draw [shift={(409,220)}, rotate = 270] [color={rgb, 255:red, 0; green, 0; blue, 0 }  ][line width=0.75]    (10.93,-3.29) .. controls (6.95,-1.4) and (3.31,-0.3) .. (0,0) .. controls (3.31,0.3) and (6.95,1.4) .. (10.93,3.29)   ;
        \draw   (108,150) .. controls (108,143.92) and (112.92,139) .. (119,139) .. controls (125.08,139) and (130,143.92) .. (130,150) .. controls (130,156.08) and (125.08,161) .. (119,161) .. controls (112.92,161) and (108,156.08) .. (108,150) -- cycle ;
        \draw    (119,161) .. controls (70,205.1) and (43.09,250.16) .. (67.44,268.89) ;
        \draw [shift={(69,270)}, rotate = 213.69] [color={rgb, 255:red, 0; green, 0; blue, 0 }  ][line width=0.75]    (10.93,-3.29) .. controls (6.95,-1.4) and (3.31,-0.3) .. (0,0) .. controls (3.31,0.3) and (6.95,1.4) .. (10.93,3.29)   ;
        \draw   (74,173) .. controls (74,167.48) and (78.48,163) .. (84,163) .. controls (89.52,163) and (94,167.48) .. (94,173) .. controls (94,178.52) and (89.52,183) .. (84,183) .. controls (78.48,183) and (74,178.52) .. (74,173) -- cycle ;
        \draw    (94,173) .. controls (167.88,213.39) and (160.25,248.92) .. (130.38,269.09) ;
        \draw [shift={(129,270)}, rotate = 327.17] [color={rgb, 255:red, 0; green, 0; blue, 0 }  ][line width=0.75]    (10.93,-3.29) .. controls (6.95,-1.4) and (3.31,-0.3) .. (0,0) .. controls (3.31,0.3) and (6.95,1.4) .. (10.93,3.29)   ;
        \draw   (227,152) .. controls (227,146.48) and (231.48,142) .. (237,142) .. controls (242.52,142) and (247,146.48) .. (247,152) .. controls (247,157.52) and (242.52,162) .. (237,162) .. controls (231.48,162) and (227,157.52) .. (227,152) -- cycle ;
        \draw    (237,162) .. controls (211.39,172.84) and (199.36,254.5) .. (228.63,269.37) ;
        \draw [shift={(230,270)}, rotate = 202.75] [color={rgb, 255:red, 0; green, 0; blue, 0 }  ][line width=0.75]    (10.93,-3.29) .. controls (6.95,-1.4) and (3.31,-0.3) .. (0,0) .. controls (3.31,0.3) and (6.95,1.4) .. (10.93,3.29)   ;
        \draw   (376,171) .. controls (376,165.48) and (380.48,161) .. (386,161) .. controls (391.52,161) and (396,165.48) .. (396,171) .. controls (396,176.52) and (391.52,181) .. (386,181) .. controls (380.48,181) and (376,176.52) .. (376,171) -- cycle ;
        \draw    (396,171) .. controls (440.78,188.91) and (499.41,234.54) .. (439.91,269.47) ;
        \draw [shift={(439,270)}, rotate = 330.15] [color={rgb, 255:red, 0; green, 0; blue, 0 }  ][line width=0.75]    (10.93,-3.29) .. controls (6.95,-1.4) and (3.31,-0.3) .. (0,0) .. controls (3.31,0.3) and (6.95,1.4) .. (10.93,3.29)   ;
        \draw    (439,250) -- (497.59,191.41) ;
        \draw [shift={(499,190)}, rotate = 135] [color={rgb, 255:red, 0; green, 0; blue, 0 }  ][line width=0.75]    (10.93,-3.29) .. controls (6.95,-1.4) and (3.31,-0.3) .. (0,0) .. controls (3.31,0.3) and (6.95,1.4) .. (10.93,3.29)   ;
        \draw  [draw opacity=0] (479,220) -- (539,220) -- (539,280) -- (479,280) -- cycle ; \draw   (509,220) -- (509,280) ; \draw   (479,250) -- (539,250) ; \draw   (479,220) -- (539,220) -- (539,280) -- (479,280) -- cycle ;
        \draw    (509,190) -- (509,218) ;
        \draw [shift={(509,220)}, rotate = 270] [color={rgb, 255:red, 0; green, 0; blue, 0 }  ][line width=0.75]    (10.93,-3.29) .. controls (6.95,-1.4) and (3.31,-0.3) .. (0,0) .. controls (3.31,0.3) and (6.95,1.4) .. (10.93,3.29)   ;
        \draw    (539,250) -- (568.11,191.79) ;
        \draw [shift={(569,190)}, rotate = 116.57] [color={rgb, 255:red, 0; green, 0; blue, 0 }  ][line width=0.75]    (10.93,-3.29) .. controls (6.95,-1.4) and (3.31,-0.3) .. (0,0) .. controls (3.31,0.3) and (6.95,1.4) .. (10.93,3.29)   ;
        \draw   (89,315) .. controls (89,306.72) and (95.72,300) .. (104,300) .. controls (112.28,300) and (119,306.72) .. (119,315) .. controls (119,323.28) and (112.28,330) .. (104,330) .. controls (95.72,330) and (89,323.28) .. (89,315) -- cycle ;
        
        \draw   (49,345) .. controls (49,336.72) and (55.72,330) .. (64,330) .. controls (72.28,330) and (79,336.72) .. (79,345) .. controls (79,353.28) and (72.28,360) .. (64,360) .. controls (55.72,360) and (49,353.28) .. (49,345) -- cycle ;
        \draw   (129,345) .. controls (129,336.72) and (135.72,330) .. (144,330) .. controls (152.28,330) and (159,336.72) .. (159,345) .. controls (159,353.28) and (152.28,360) .. (144,360) .. controls (135.72,360) and (129,353.28) .. (129,345) -- cycle ;
        \draw    (89,315) .. controls (84.2,313.08) and (69.26,314.85) .. (64.54,328.26) ;
        \draw [shift={(64,330)}, rotate = 284.93] [color={rgb, 255:red, 0; green, 0; blue, 0 }  ][line width=0.75]    (10.93,-3.29) .. controls (6.95,-1.4) and (3.31,-0.3) .. (0,0) .. controls (3.31,0.3) and (6.95,1.4) .. (10.93,3.29)   ;
        \draw    (79,345) .. controls (99.58,345) and (100.95,345) .. (127.35,345) ;
        \draw [shift={(129,345)}, rotate = 180] [color={rgb, 255:red, 0; green, 0; blue, 0 }  ][line width=0.75]    (10.93,-3.29) .. controls (6.95,-1.4) and (3.31,-0.3) .. (0,0) .. controls (3.31,0.3) and (6.95,1.4) .. (10.93,3.29)   ;
        \draw   (249,315) .. controls (249,306.72) and (255.72,300) .. (264,300) .. controls (272.28,300) and (279,306.72) .. (279,315) .. controls (279,323.28) and (272.28,330) .. (264,330) .. controls (255.72,330) and (249,323.28) .. (249,315) -- cycle ;
        
        \draw   (209,345) .. controls (209,336.72) and (215.72,330) .. (224,330) .. controls (232.28,330) and (239,336.72) .. (239,345) .. controls (239,353.28) and (232.28,360) .. (224,360) .. controls (215.72,360) and (209,353.28) .. (209,345) -- cycle ;
        \draw   (289,345) .. controls (289,336.72) and (295.72,330) .. (304,330) .. controls (312.28,330) and (319,336.72) .. (319,345) .. controls (319,353.28) and (312.28,360) .. (304,360) .. controls (295.72,360) and (289,353.28) .. (289,345) -- cycle ;
        \draw   (249,380) .. controls (249,371.72) and (255.72,365) .. (264,365) .. controls (272.28,365) and (279,371.72) .. (279,380) .. controls (279,388.28) and (272.28,395) .. (264,395) .. controls (255.72,395) and (249,388.28) .. (249,380) -- cycle ;
        \draw    (289,345) .. controls (281.2,347.93) and (264.84,340.39) .. (264.03,363.19) ;
        \draw [shift={(264,365)}, rotate = 270] [color={rgb, 255:red, 0; green, 0; blue, 0 }  ][line width=0.75]    (10.93,-3.29) .. controls (6.95,-1.4) and (3.31,-0.3) .. (0,0) .. controls (3.31,0.3) and (6.95,1.4) .. (10.93,3.29)   ;
        \draw    (249,315) .. controls (234.45,315) and (227.43,310.3) .. (224.28,328.27) ;
        \draw [shift={(224,330)}, rotate = 278.53] [color={rgb, 255:red, 0; green, 0; blue, 0 }  ][line width=0.75]    (10.93,-3.29) .. controls (6.95,-1.4) and (3.31,-0.3) .. (0,0) .. controls (3.31,0.3) and (6.95,1.4) .. (10.93,3.29)   ;
        \draw   (359,345) .. controls (359,336.72) and (365.72,330) .. (374,330) .. controls (382.28,330) and (389,336.72) .. (389,345) .. controls (389,353.28) and (382.28,360) .. (374,360) .. controls (365.72,360) and (359,353.28) .. (359,345) -- cycle ;
        \draw   (439,345) .. controls (439,336.72) and (445.72,330) .. (454,330) .. controls (462.28,330) and (469,336.72) .. (469,345) .. controls (469,353.28) and (462.28,360) .. (454,360) .. controls (445.72,360) and (439,353.28) .. (439,345) -- cycle ;
        \draw   (399,380) .. controls (399,371.72) and (405.72,365) .. (414,365) .. controls (422.28,365) and (429,371.72) .. (429,380) .. controls (429,388.28) and (422.28,395) .. (414,395) .. controls (405.72,395) and (399,388.28) .. (399,380) -- cycle ;
        \draw    (389,345) .. controls (402.72,345) and (406.84,345.96) .. (437.11,345.06) ;
        \draw [shift={(439,345)}, rotate = 178.21] [color={rgb, 255:red, 0; green, 0; blue, 0 }  ][line width=0.75]    (10.93,-3.29) .. controls (6.95,-1.4) and (3.31,-0.3) .. (0,0) .. controls (3.31,0.3) and (6.95,1.4) .. (10.93,3.29)   ;
        \draw   (569,320) .. controls (569,311.72) and (575.72,305) .. (584,305) .. controls (592.28,305) and (599,311.72) .. (599,320) .. controls (599,328.28) and (592.28,335) .. (584,335) .. controls (575.72,335) and (569,328.28) .. (569,320) -- cycle ;
        \draw   (609,355) .. controls (609,346.72) and (615.72,340) .. (624,340) .. controls (632.28,340) and (639,346.72) .. (639,355) .. controls (639,363.28) and (632.28,370) .. (624,370) .. controls (615.72,370) and (609,363.28) .. (609,355) -- cycle ;
        \draw   (569,395) .. controls (569,386.72) and (575.72,380) .. (584,380) .. controls (592.28,380) and (599,386.72) .. (599,395) .. controls (599,403.28) and (592.28,410) .. (584,410) .. controls (575.72,410) and (569,403.28) .. (569,395) -- cycle ;
        
        \draw    (599,320) .. controls (614.52,325.82) and (617.81,318.47) .. (623.47,338.1) ;
        \draw [shift={(624,340)}, rotate = 254.74] [color={rgb, 255:red, 0; green, 0; blue, 0 }  ][line width=0.75]    (10.93,-3.29) .. controls (6.95,-1.4) and (3.31,-0.3) .. (0,0) .. controls (3.31,0.3) and (6.95,1.4) .. (10.93,3.29)   ;
        \draw    (624,370) .. controls (608.4,375.85) and (605.16,366.49) .. (585.54,378.99) ;
        \draw [shift={(584,380)}, rotate = 326.31] [color={rgb, 255:red, 0; green, 0; blue, 0 }  ][line width=0.75]    (10.93,-3.29) .. controls (6.95,-1.4) and (3.31,-0.3) .. (0,0) .. controls (3.31,0.3) and (6.95,1.4) .. (10.93,3.29)   ;
        \draw    (454,360) .. controls (450.12,372.61) and (448.12,372.05) .. (430.66,379.3) ;
        \draw [shift={(429,380)}, rotate = 337.17] [color={rgb, 255:red, 0; green, 0; blue, 0 }  ][line width=0.75]    (10.93,-3.29) .. controls (6.95,-1.4) and (3.31,-0.3) .. (0,0) .. controls (3.31,0.3) and (6.95,1.4) .. (10.93,3.29)   ;
        \draw   (529,355) .. controls (529,346.72) and (535.72,340) .. (544,340) .. controls (552.28,340) and (559,346.72) .. (559,355) .. controls (559,363.28) and (552.28,370) .. (544,370) .. controls (535.72,370) and (529,363.28) .. (529,355) -- cycle ;
        \draw    (544,340) .. controls (545.92,328.48) and (553.37,324.33) .. (567.23,320.48) ;
        \draw [shift={(569,320)}, rotate = 165.07] [color={rgb, 255:red, 0; green, 0; blue, 0 }  ][line width=0.75]    (10.93,-3.29) .. controls (6.95,-1.4) and (3.31,-0.3) .. (0,0) .. controls (3.31,0.3) and (6.95,1.4) .. (10.93,3.29)   ;
        
        \draw (484,372.4) node [anchor=north west][inner sep=0.75pt]    {$\cdots $};
        \draw (205,42.4) node [anchor=north west][inner sep=0.75pt]    {$\langle a_{\lozenge } ,\ b_{\lozenge } ,\ a_{\square } ,\ c_{\lozenge } ,\ b_{\square } ,\ d_{\lozenge } ,\ a_{\square } ,\ e_{\square } ,\ c_{\lozenge } ,\ e_{\lozenge } \rangle $};
        \draw (341,82.4) node [anchor=north west][inner sep=0.75pt]    {$\cdots $};
        \draw (484,152.4) node [anchor=north west][inner sep=0.75pt]    {$\cdots $};
        \draw (530,163.4) node [anchor=north west][inner sep=0.75pt]    {$e_{\square }\rightarrow E$};
        \draw (530,141.4) node [anchor=north west][inner sep=0.75pt]    {$\textcolor[rgb]{0.74,0.06,0.88}{d}\textcolor[rgb]{0.74,0.06,0.88}{_{\lozenge }}\textcolor[rgb]{0.74,0.06,0.88}{\rightarrow }\textcolor[rgb]{0.74,0.06,0.88}{c}\textcolor[rgb]{0.74,0.06,0.88}{_{\lozenge }}\textcolor[rgb]{0.74,0.06,0.88}{\rightarrow }\textcolor[rgb]{0.74,0.06,0.88}{e}\textcolor[rgb]{0.74,0.06,0.88}{_{\lozenge }}\rightarrow E$};
        \draw (239,92.4) node [anchor=north west][inner sep=0.75pt]    {$\textcolor[rgb]{0.82,0.01,0.11}{c}\textcolor[rgb]{0.82,0.01,0.11}{_{\lozenge }}$};
        \draw (300,102.4) node [anchor=north west][inner sep=0.75pt]    {$\textcolor[rgb]{0.96,0.65,0.14}{b}\textcolor[rgb]{0.96,0.65,0.14}{_{\square }}$};
        \draw (48,45) node [anchor=north west][inner sep=0.75pt]   [align=left] {ID};
        \draw (9,77) node [anchor=north west][inner sep=0.75pt]   [align=left] {Last EC};
        \draw (530,82.4) node [anchor=north west][inner sep=0.75pt]    {$\textcolor[rgb]{0.74,0.06,0.88}{e}\textcolor[rgb]{0.74,0.06,0.88}{_{\lozenge }}$};
        \draw (139,337.4) node [anchor=north west][inner sep=0.75pt]    {$b$};
        \draw (59,335.4) node [anchor=north west][inner sep=0.75pt]    {$a$};
        \draw (98,306.4) node [anchor=north west][inner sep=0.75pt]    {$S$};
        \draw (81,255.4) node [anchor=north west][inner sep=0.75pt]    {$\textcolor[rgb]{0.82,0.01,0.11}{b}\textcolor[rgb]{0.82,0.01,0.11}{}$};
        \draw (110,259.4) node [anchor=north west][inner sep=0.75pt]    {$a$};
        \draw (79,226.4) node [anchor=north west][inner sep=0.75pt]    {$\lozenge $};
        \draw (106,226.4) node [anchor=north west][inner sep=0.75pt]    {$\square $};
        \draw (240,259.4) node [anchor=north west][inner sep=0.75pt]    {$c$};
        \draw (269,259.4) node [anchor=north west][inner sep=0.75pt]    {$\textcolor[rgb]{0.96,0.65,0.14}{a}$};
        \draw (238,227.4) node [anchor=north west][inner sep=0.75pt]    {$\lozenge $};
        \draw (267,227.4) node [anchor=north west][inner sep=0.75pt]    {$\square $};
        \draw (351,160.4) node [anchor=north west][inner sep=0.75pt]    {$\textcolor[rgb]{0.96,0.65,0.14}{a}\textcolor[rgb]{0.96,0.65,0.14}{_{\square }}\textcolor[rgb]{0.96,0.65,0.14}{\rightarrow }\textcolor[rgb]{0.96,0.65,0.14}{b}\textcolor[rgb]{0.96,0.65,0.14}{_{\square }}$};
        \draw (351,140.4) node [anchor=north west][inner sep=0.75pt]    {$b_{\lozenge }\rightarrow c_{\lozenge }$};
        \draw (390,259.4) node [anchor=north west][inner sep=0.75pt]    {$\textcolor[rgb]{0.49,0.83,0.13}{c}$};
        \draw (419,256.4) node [anchor=north west][inner sep=0.75pt]    {$b$};
        \draw (388,227.4) node [anchor=north west][inner sep=0.75pt]    {$\lozenge $};
        \draw (416,228.4) node [anchor=north west][inner sep=0.75pt]    {$\square $};
        \draw (490,259.4) node [anchor=north west][inner sep=0.75pt]    {$\textcolor[rgb]{0.74,0.06,0.88}{c}$};
        \draw (489,227.4) node [anchor=north west][inner sep=0.75pt]    {$\lozenge $};
        \draw (75,76.4) node [anchor=north west][inner sep=0.75pt]  [color={rgb, 255:red, 0; green, 0; blue, 0 }  ,opacity=1 ]  {$\emptyset $};
        \draw (105,76.4) node [anchor=north west][inner sep=0.75pt]  [color={rgb, 255:red, 0; green, 0; blue, 0 }  ,opacity=1 ]  {$\emptyset $};
        \draw (76,46.4) node [anchor=north west][inner sep=0.75pt]  [color={rgb, 255:red, 0; green, 0; blue, 0 }  ,opacity=1 ]  {$\emptyset $};
        \draw (105,45.4) node [anchor=north west][inner sep=0.75pt]  [color={rgb, 255:red, 0; green, 0; blue, 0 }  ,opacity=1 ]  {$\emptyset $};
        \draw (517,225.4) node [anchor=north west][inner sep=0.75pt]  [color={rgb, 255:red, 0; green, 0; blue, 0 }  ,opacity=1 ]  {$\emptyset $};
        \draw (517,257.4) node [anchor=north west][inner sep=0.75pt]  [color={rgb, 255:red, 0; green, 0; blue, 0 }  ,opacity=1 ]  {$\emptyset $};
        \draw (299,337.4) node [anchor=north west][inner sep=0.75pt]    {$b$};
        \draw (219,335.4) node [anchor=north west][inner sep=0.75pt]    {$a$};
        \draw (260,370.4) node [anchor=north west][inner sep=0.75pt]    {$c$};
        \draw (258,306.4) node [anchor=north west][inner sep=0.75pt]    {$S$};
        \draw (449,337.4) node [anchor=north west][inner sep=0.75pt]    {$b$};
        \draw (369,335.4) node [anchor=north west][inner sep=0.75pt]    {$a$};
        \draw (410,370.4) node [anchor=north west][inner sep=0.75pt]    {$c$};
        \draw (580,310.4) node [anchor=north west][inner sep=0.75pt]    {$c$};
        \draw (620,345.4) node [anchor=north west][inner sep=0.75pt]    {$e$};
        \draw (578,386.4) node [anchor=north west][inner sep=0.75pt]    {$E$};
        \draw (539,346.4) node [anchor=north west][inner sep=0.75pt]    {$d$};
        \draw (49,140.4) node [anchor=north west][inner sep=0.75pt]    {$S\rightarrow a_{\lozenge }\rightarrow b_{\lozenge }$};
        \draw (47,162.4) node [anchor=north west][inner sep=0.75pt]    {$S\rightarrow a_{\square }$};
        \draw (201,140.4) node [anchor=north west][inner sep=0.75pt]    {$\textcolor[rgb]{0.82,0.01,0.11}{b}\textcolor[rgb]{0.82,0.01,0.11}{_{\lozenge }}\textcolor[rgb]{0.82,0.01,0.11}{\rightarrow }\textcolor[rgb]{0.82,0.01,0.11}{c}\textcolor[rgb]{0.82,0.01,0.11}{_{\lozenge }}$};
        \draw (200,162.4) node [anchor=north west][inner sep=0.75pt]    {$S\rightarrow a_{\square }$};
        \draw (167,92) node [anchor=north west][inner sep=0.75pt]   [align=left] {Init};

        \end{tikzpicture}

        }
        \caption{Generation of features using TET. $S$ represents the SOT (start of trace) token, while $E$ represents the EOT (end of trace) token. The frequent occurrence of an event class within the window causes the weights of nodes and edges to increase; however, the weights are not reflected in this diagram.}
        \label{fig:online_process_mining_with_tet}
    \end{figure*}
    
    Here we add a more detailed, accurate and general version of the algorithm applicable to various types of streams based on Zhong et al.'s algorithm \cite{zhong2022process}.
    
    Generally, we attempt to utilise the global flow discovery capability of process mining, but design it online. As mentioned previously, process mining analyses the relationships between packets in flows and encodes the global flow structure into the process model rather than analysing the flows themselves. The resulting algorithm will be a transition-based preprocessor that takes streams of activities as the input and produces a series of adjacency/transition matrices. These matrices have historical information encoded.
    
    Before discussing the algorithm, it is necessary to re-define the concepts of transitions and event classes, whose traditional definitions in process mining may vary slightly. 
    
    Given a sequence of events $P$, we define a \emph{transition} in $P$ as a pair of consecutive events $\left(p_{i},p_{j}\right)$ within the same case. The transition is referred to as the precedence relation, and $P$ can be treated as the event log.
        
    A \emph{trace} is a series of event names. A \emph{case} is a trace instance that has been executed. A trace, for example, is a predefined procedure/process for producing a certain type of medication. This type of medication can be manufactured multiple times, resulting in numerous cases. The event log is comprised of the logs from the production of various types of medications. The case is also referred to as a flow in this paper.
    
    Here is an example, giving a series of events $P = \langle p_1, p_2,p_3, p_4, p_5 \rangle$ with two flows $t_1$ and $t_2$, where flow $t_1 = \langle p_1,p_3,p_5\rangle $ and flow $t_2 = \langle p_2,p_4\rangle$, we will get two transitions for $t_1$: $\left(p_1,p_3\right)$ and $\left(p_3,p_5\right)$; one transition for $t_2$: $\left(p_2,p_4\right)$. $p_1$ and $p_2$ are two consecutive events, however, these events will not be considered as a transition as they belong to different flows.
    
    An \emph{event class} $ec(p)$ is the name of an event $p$, and normally it is the concatenated string of non-numerical attribute data. In the example of network traffic, it is the enabled flags of packets and whether the packet comes from the server of the client \cite{zhong2022process}. An event is the executed instance of an event class.
    
    The temporal event table (TET), formerly known as the state table in \cite{zhong2022process}, is a data structure used to prevent the loss of information on transitions for historical data not covered by the sliding window. We call it TET in this paper to eliminate the confusion. Let us consider a smaller scaled example which has an event log of 10 events that involved in 2 concurrent traces where trace $\lozenge = \langle a_\lozenge, b_\lozenge, c_\lozenge, d_\lozenge, c_\lozenge, e_\lozenge \rangle$ and trace $\square = \langle a_\square, b_\square, a_\square, e_\square \rangle$; we have 5 event classes $\{ a, b, c, d, e\}$ and assuming our sliding window size $l$ is 3.
    
    The TET has a header, which is typically the case ID, and the event class is stored beneath the case ID. In Figure \ref{fig:online_process_mining_with_tet}, the TET is initialised with the initial window and the Last EC value is updated based on the most recent packet. For each incoming packet, the system uses the case ID from the incoming packet as a key to retrieve the previous event. For instance, $c_\lozenge$ has case $\lozenge$, which exists in TET, and the event class record for $\lozenge$ in TET is $b$; the system creates the relation $(b, c)$ and then updates the record in TET to the last observation $c$. When the traces reach their end point, the records will be null and the associated key/ID in the TET will be removed. TET works in conjunction with the sliding window and the resulting output are transition matrices with size $n^2$ where $n$ is the number of observed event classes.
    
    TET in combination with a sliding window buffer that retains the previous $l$ transitions eliminates the need to generate a graph for each window and the need to search for the previous event with the same trace ID as the incoming event. TET stores the event history and maintains an $O(1)$ computational complexity for processing each activity. TET can be expanded to support variable-length historical events logging, and sliding windows are compatible with process mining techniques such as trace clustering and abstraction \cite{gunther2007fuzzy,song2008trace}.
    
    The main difference between TFGen and traditional process mining is that process mining focus on process model generation and analytical method like conformance checking on process models, whereas TFGen is designed for online processing and feature generation for machine learning. The TFGen implementation as a Python package can be found on Github\footnote{https://github.com/yinzheng-zhong/TFGen}. This implementation is capable of processing around 80,000 events per second using a single thread of an Intel Core i5-12600K processor\footnote{Using the provided NIDS dataset on Github}.

\section{Discussion}
    This novel feature generator was originally developed for NIDS and the report on its performance can be found in \cite{zhong2022process}.
    Some results presented in \cite{zhong2022process} shows AUC under 0.5 and we believe this is due to the fact that some transition frequencies stabilise under attack. Therefore, the attack data have lower variance and may be characterised as attacks by some outlier detectors.
    
    Further improvements obviously can be done. Given the generality and flexibility of the algorithm, the position for our paper is that TFGen may be applicable to the domain of transition-based problems or data that can be mapped to discrete space. Here are some instances.
    
    \begin{itemize}
        \item It is applicable to HIDS based on system calls or kernel operations \cite{liu2018host,byrnes2020modern,kadar2019system}. The calls are provided as events, and each process will generate a distinct case. With a larger scaled environment, agents can be deployed on multiple systems for concurrent data collection from a cluster of hosts; TFGen is ideally applicable as long as cases can be modelled and the lengths of cases are finite.
        
        \item Computer vision and sensor-based security systems \cite{DING2018118,luo2018computer} that detect and monitor a series of activities for health and safety measurement. Instead of using the current approaches, the series of classified activities can be modelled as cases where each case can be produced by specific personnel. Each case consists of events that have several attributes such as gesture, department and gender. The benefit could be better overall performance, and the behaviours of multiple personnel are encoded.
        \item Anomaly detection in the operation of critical infrastructure \cite{gauthama2019anomaly}, where TFGen can be used in conjunction with numerical sensor readings to encode time-series activity data.
    \end{itemize}
    
    These applications are based on hypotheses that TFGen supports all standard event logs as long as processes can be converted to event log, and the performance and practicability of using TFGen in these areas can be open research questions for the future work.
    
    To demonstrate this, we conduct a quick experiment to evaluate the performance of HIDS using the dataset of API calls captured by Cuckoo Sandbox \cite{matthew_nunes_2018_1203289,nunes2019getting}. Since the dataset does not include a native event log, event logs are generated based on the timestamps and API names. Events are extracted from logs of all processes, then combined and sorted based on the timestamps.
    
    To reduce the dimensionality, we only generate transition matrices based on a limited number of most frequent event classes out of over 260 observable event classes. We call the event classes that fall within the limited range visible event classes, and we call other event classes hidden event classes. Hidden event classes are counted into the default event class "Other" for frequency calculation. the setups are available below.
    
    \begin{itemize}
        \item t100\_ipca0: Limiting the visible event classes to 100.
        \item t50\_ipca0: Limiting the visible event classes to 50.
        \item t25\_ipca0: Limiting the visible event classes to 25.
        \item t10\_ipca0: Limiting the visible event classes to 10.
    \end{itemize}
    
\begin{figure}[]
        \centering
        \includegraphics[width=3in]{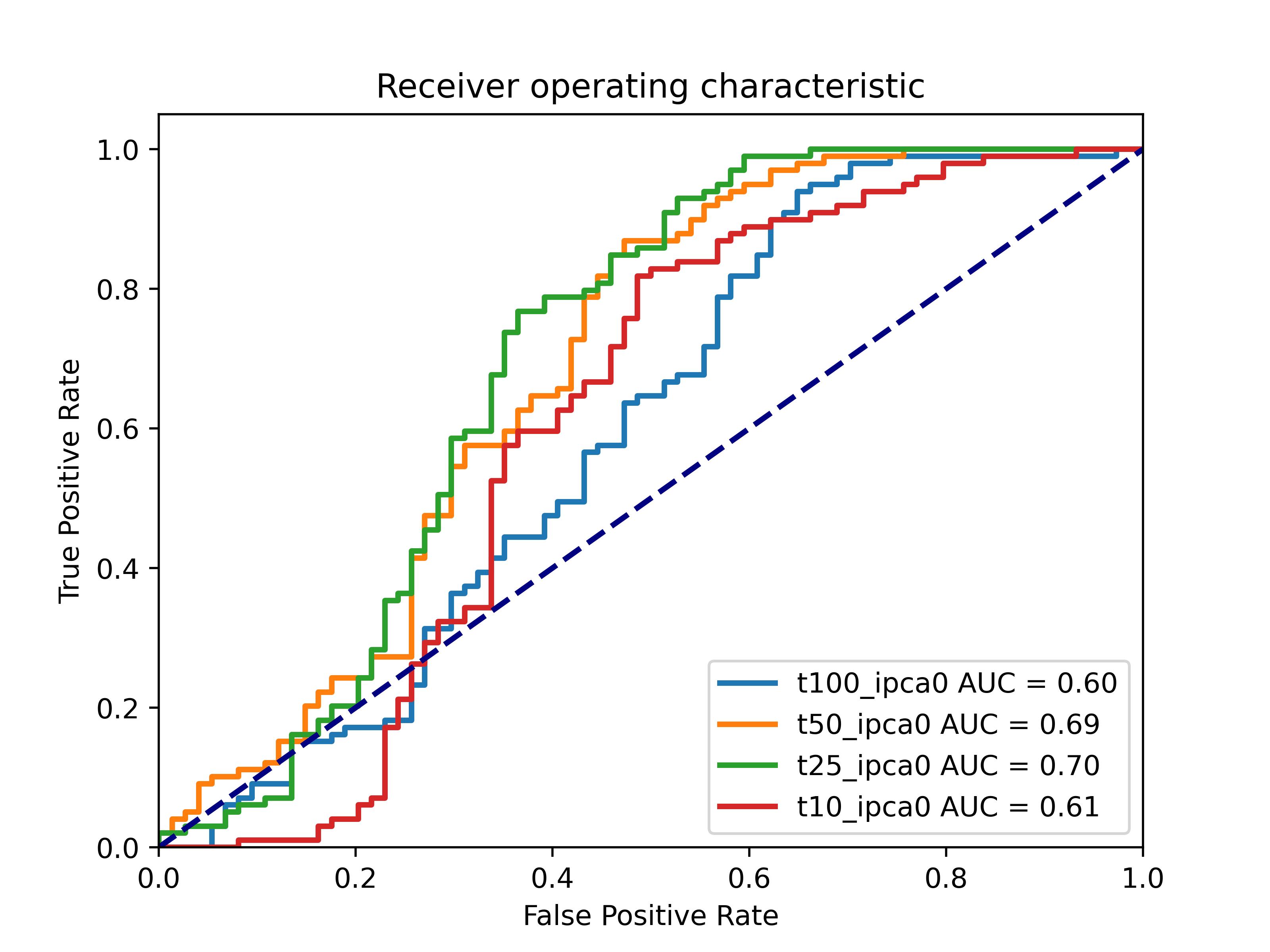}
        \caption{Performance using Convolutional Autoencoder.}
        \label{fig:roc}
    \end{figure}

    Figure \ref{fig:roc} is the result of utilising the Convolutional Autoencoder for unsupervised learning on data generated by TFGen. The outlier factor of a case is determined by the event with the highest outlier factor. The Convolutional Autoencoder is able to inference over 28,000 events/s using batch size 32 on a single RTX 2070 graphics cards. Details of the constructed event logs and the implementation can be accessed on Zenodo\footnote{https://doi.org/10.5281/zenodo.7154396}. The link also provides a document that shows extensive details of this experiment.
    
    Further analysis and experimentation will be conducted, and a better dataset containing native event log data may be utilised. In contrast to a native event log, this generated log may not provide a true representation of concurrent processes.
    
    These are disadvantages of TFGen, for instance, a smaller window size may result in sparse matrices, necessitating an adjustment of the window size based on the problem and anomaly detector. When event classes contain numerous attributes, such as the HIDS dataset we used previously, the output matrices may be of high dimension. Using dimension reduction techniques such as incremental principal component analysis (IPCA) or a list of limited event classes is possible.
    
    To clarify, the purpose of this research is not to demonstrate the high accuracy and low FPR of any experiment, but rather to demonstrate that the generalised approach has the potential to function in other domains. Due to the fact that a high FPR is associated with the poor practicality of AIDS in general, this algorithm may provide an alternative strategy.




\bibliographystyle{IEEEtran}
\bibliography{ref}

\end{document}